\documentclass[11pt]{article}

\usepackage{geometry}
\usepackage{authblk}
\usepackage[backend=biber,sorting=none,style=nature]{biblatex}

\usepackage{amsmath}
\usepackage{amssymb}
\usepackage{bm}
\usepackage{booktabs}
\usepackage{float}
\usepackage{graphicx}

\usepackage[parfill]{parskip}
\usepackage{geometry}
\usepackage{hyperref}
\usepackage{xcolor}

\hypersetup{
    colorlinks  = true,
    urlcolor    = black!32!blue,
    linkcolor   = black!32!blue,
    citecolor   = black!32!blue,
    pdfauthor   = {David Oniani, Jordan Hilsman, Hang Dong, Shiven Verma, Fengyi Gao, Yanshan Wang},
    pdftitle    = {\Large Large Language Models Vote: Prompting for Rare Disease Identification},
    pdfkeywords = {artificial intelligence, computer science, natural language processing, generative artificial intelligence},
    pdflang     = English,
    pdfpagemode = UseNone,
    pdfsubject  = {\Large Large Language Models Vote: Prompting for Rare Disease Identification}
}

\let\emptyset\varnothing

\DeclareMathOperator*{\argmax}{arg\,max}

\DeclareSymbolFont{extraup}{U}{zavm}{m}{n}
\DeclareMathSymbol{\varheart}{\mathalpha}{extraup}{86}
\DeclareMathSymbol{\vardiamond}{\mathalpha}{extraup}{87}
\newcommand{\h}[1]{\underline{\textbf{#1}}}

\input{glyphtounicode}
\title{\Large Large Language Models Vote: Prompting for Rare Disease Identification}

\author[1]{David Oniani}
\author[1]{Jordan Hilsman}
\author[2]{Hang Dong}
\author[3]{Shiven Verma}
\author[1]{\authorcr Fengyi Gao}
\author[1]{Yanshan Wang}

\affil[1]{\small Department of Health Information Management, University of Pittsburgh, Pittsburgh, PA, USA}
\affil[2]{\small Department of Computer Science, University of Oxford, Oxford, UK}
\affil[3]{\small Upper St. Clair HS, USA}

\date{}

\begin{document}

\maketitle


\begin{abstract}
  \noindent The emergence of generative Large Language Models (LLMs) emphasizes the need for
  accurate and efficient prompting approaches. The use of LLMs in Few-Shot Learning (FSL) settings,
  where data is scarce, has become a standard practice. FSL has also become popular in many
  Artificial Intelligence (AI) subdomains, including AI for health. Rare diseases affect a small
  fraction of the population, and due to limited data availability, their identification from
  clinical notes inherently requires FSL techniques. Manual data collection and annotation is both
  expensive and time-consuming. In this paper, we propose Models-Vote Prompting (MVP), an ensemble
  prompting approach for improving the performance of LLM queries in FSL settings. MVP works by
  prompting several LLMs to perform the same task and then conducting a majority vote on the
  resulting outputs. The proposed method achieves improved results to any one model in the ensemble
  on one-shot rare disease identification and classification tasks. MVP performance rivals that of
  well-known Self-Consistency (SC) prompting. In addition, we introduce a novel rare disease dataset
  for FSL, reproducible to those who signed the MIMIC-IV Data Use Agreement (DUA). Furthermore, we
  also assess the feasibility of using JSON-augmented prompts for automating generative LLM
  evaluation.
\end{abstract}


\section*{Introduction}

Large Language Models (LLMs) are language models that typically consist of hundreds of millions to
billions of parameters and are characterized by unsupervised pre-training and supervised
fine-tuning~\cite{llmsurvey}. LLMs have proven useful in many tasks, including representation
learning~\cite{bert, roberta}, machine translation~\cite{bart, t5}, and text generation~\cite{gpt-2,
gpt-3}. Recently, generative LLMs have taken Computer Science (CS) and Artificial Intelligence (AI)
research communities by storm, with models such as LLaMA~\cite{llama} and Stable
Diffusion~\cite{stablediffusion} seeing significant adoption and with ChatGPT~\cite{chatgpt}
reaching over 100 million active users within two months since its release~\cite{chatgpt100mm}.

Prompting is a novel paradigm, where with the help of a textual prompt, downstream tasks can be
modeled similar to those solved during pre-training~\cite{ppp}. This can be considered an
alternative to the conventional unsupervised pre-training and supervised fine-tuning paradigms. The
act of prompting is closely tied to the concept of in-context learning, where a language model is
given a prompt composed of training examples and a test instance as the input. The model then
generates the corresponding output to the test instance without any change to its
parameters~\cite{rubin_icllearning}. The output of the model coherently follows the language of the
prompt, by understanding the meaning present in the examples~\cite{jo_llm_health}. Prompting led to
the emergence of prompt engineering, a discipline that aims to develop effective prompting methods
for efficiently solving tasks~\cite{promptengineering}. A variety of prompting methods have already
been developed, including Instruction Prompting (IP) in the manner of
InstructGPT~\cite{instructgpt}, Chain-of-Thought (CoT)~\cite{cot}, and Self-Consistency
(SC)~\cite{sc} prompting. Yet, there has been a lack of prompting approaches that combine the
knowledge of multiple LLMs.

Deep Learning (DL) often requires large amounts of data that can be expensive and occasionally
difficult to obtain. Few-Shot Learning (FSL) is a subfield of AI that attempts to enable machine
learning even in cases with a small number sample (also known as shots). FSL has recently shown
promising performance in various tasks, from image segmentation~\cite{fslis} to speaker
recognition~\cite{fslsr} and from Named-Entity Recognition (NER)~\cite{fslner} to Question-Answering
(QA)~\cite{fslqa}. Furthermore, prompt-based approaches perform well in FSL
settings~\cite{fslandprompting}.

Rare disease identification serves as a natural application for utilizing FSL techniques. A rare
disease is defined as a disease that affects no more than 200,000 people in the population (US
definition)~\cite{USCode21sect360bb} or no more than one in two thousand people (EU
definition)~\cite{EU}. Despite slight differences in these definitions, we can consider a rare
disease to be a disease that affects one in several thousand people. As a result of this rarity, it
is hard to obtain extensive and comprehensive amounts of data for these diseases, which prompts the
use of FSL approaches. Unlike structured Electronic Health Records (EHRs) that primarily capture
standardized and limited information, Clinical Notes (CNs) are detailed narratives of patient
conditions, symptoms, treatments, and contextual information. The complexity and variety of language
used in CNs reflect the intricate nuances of medical cases, including subtle symptoms that might not
be documented in structured formats. NLP algorithms can effectively sift through these notes,
extracting valuable insights and patterns that might lead to the early detection and accurate
diagnosis of rare diseases. This process aids medical professionals in uncovering hidden
correlations and symptoms, reducing the diagnostic odyssey that often characterizes rare disease
cases. On the other hand, rare disease datasets are also scarce, especially for Natural Language
Processing (NLP) purposes, and relying only on the structured data can lead to erroneous conclusions
or missed patients when recruiting for trials~\cite{Ford2013, HernandezBoussard2016, Kharrazi2018}.
To the best of our knowledge, there has been one work that attempted to build a rare disease dataset
via weak supervision~\cite{rdws}, but the annotations have not been fully verified by humans with
biomedical experience, and the number of rare disease cases in existing datasets such as
MIMIC-III~\cite{mimic-iii, mimic-iii-original} are not sufficient for 4-way 128 or 256-shot
experiments.

In this paper, we make the following contributions:

\begin{enumerate}
    \item We propose Models-Vote Prompting (MVP), an ensemble prompting method that uses multiple
          LLMs for majority voting to improve the success rate of task completion. We evaluate the
          proposed strategy on two NLP tasks (with four different context sizes) and measure its
          performance via metrics: accuracy, precision, recall, and F-score. We also compare MVP to
          SC prompting. Furthermore, we conduct statistical hypothesis testing to verify that the
          mean difference in model outputs between MVP and second-best models is non-zero.
    \item We introduce a novel FSL dataset for rare disease identification. The rare disease dataset
          was obtained by processing a recently released MIMIC-IV~\cite{mimic-iv, mimic-iv-original}
          database. For the tasks performed in the study, we conduct a thorough, two-round
          annotation process involving annotation guidelines and two annotators with biomedical
          experience, resulting in a high Inter-Annotator Agreement (IAA) score. The dataset,
          excluding annotations, can be fully reproduced by following the steps in the codebase we
          released for the study\footnote{\url{https://github.com/PittNAIL/llms-vote}.}.
    \item We emphasize the importance of incorporating parsable formats, such as JSON, in LLM
          prompts, for facilitating LLM evaluation.
\end{enumerate}


\section*{Overview of Prompting Approaches}

In this section, we provide a brief overview of the popular prompting approaches: IP, CoT, and SC.
We also formally define our proposed approach: MVP. In this direction, we use the formal approach
inspired by Phuong et al.~\cite{phuong2022formal} We start by introducing the formal notation and
defining the existing methods. A comparison with MVP is made in the following section.

\subsection*{Notation}

Let \(\boldsymbol{x} \equiv x[1:n] \equiv x[1], x[2], \dots, x[n]\) be a sequence of sub-word
tokens. The goal is to learn an estimate \(\hat{P}\) of the distribution \(P(x)\) from independent
and identically distributed (i.i.d.) data. We denote the parameters of the distribution as
\(\theta\) and the output sequence of tokens as \(\boldsymbol{y}\). Note that, unlike a typical
programming language, we use 1-based indexing and inclusive ranges (i.e., range \(x[i : j]\)
includes both \(x[i]\) and \(x[j]\)). Then the following holds:
\begin{align}
  \begin{split}
    \hat{P}(\boldsymbol{x}) = P_\theta(x[1]) &\cdot P_\theta(x[2] \mid x[1])\\
                                             &\cdot P_\theta(x[3] \mid x[1:2])\\
                                             &\cdot P_\theta(x[4] \mid x[1:3])\\
                                             &\cdots\\
                                             &\cdot P_\theta(x[n] \mid x[1:n - 1])
  \end{split}
\end{align}
which can also be formulated as
\begin{equation}
  \hat{P}(\boldsymbol{x}) = P_\theta(x[1]) \cdot \prod_{i = 2}^n P_\theta(x[i] \mid x[1:i-1])
\end{equation}

\subsection*{Overview}

IP is the simplest of all the prompting approaches described in the paper, where the input text
includes instructions (usually, a few). This can be formally expressed as follows:

\begin{equation}
  \boldsymbol{y} \sim P_\theta^{\text{IP}}(\boldsymbol{y} \mid \boldsymbol{x})
\end{equation}

CoT makes use of a series of reasoning steps \(\boldsymbol{s} \equiv s_1, s_2, \dots s_l\) to get
from \(\boldsymbol{x}\) to the output \(\boldsymbol{y}\). Hence, it can be formulated as:
\begin{equation}
  \boldsymbol{y} \sim P_\theta^{\text{CoT}}(\boldsymbol{y} \mid \boldsymbol{x}, \boldsymbol{s})
\end{equation}

SC prompting is an ensemble approach that samples LLM decoder to generate some number of i.i.d
chains of thought. We denote this number as \(m\) and let \(\boldsymbol{c} \equiv c_1, c_2, \dots
c_m\) be the chains of thought. We get a set of responses \(\boldsymbol{r} \equiv r_1, r_2, \dots
r_m\), where \(r_i = P_\theta^{CoT}(\boldsymbol{y} | c_i)\). After generating \(\boldsymbol{r}\), we
get:
\begin{equation}
  \boldsymbol{y} \sim \#\argmax{\{i \mid r_i = \boldsymbol{y}\}}
\end{equation}

\begin{figure*}[t!]
  \begin{center}
    \includegraphics[width=\linewidth]{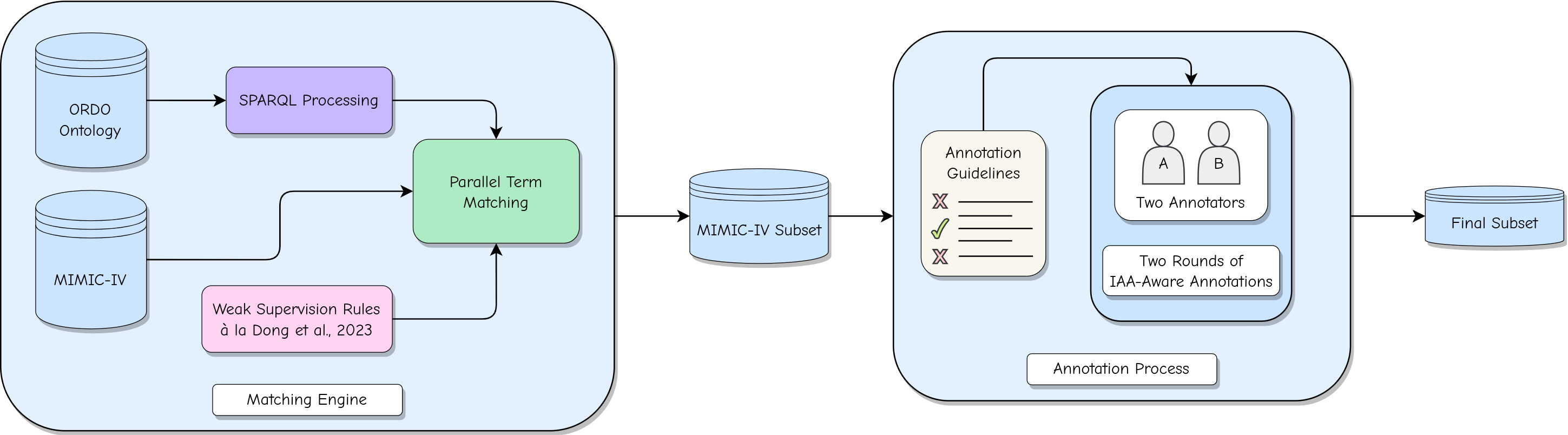}
  \end{center}
  \caption{Rare Disease Dataset Pipeline}
  \label{fig:dataset_pipeline}
\end{figure*}


\section*{Models-Vote Prompting}

Notice that all of the methods described in the previous section make use of a single model, yet the
availability of LLMs allows for the use of multiple, further enriching model pool, and diversity of
the training data. MVP is an ensemble approach that uses a set of \(m\) language models to generate
a response. Formally, this can be expressed as follows:
\begin{align}
  \begin{split}
    \boldsymbol{y_1} &\sim P^{MVP}_{\theta(1)}(\boldsymbol{y} \mid \boldsymbol{x})\\
    \boldsymbol{y_2} &\sim P^{MVP}_{\theta(2)}(\boldsymbol{y} \mid \boldsymbol{x})\\
    \dots\\
    \boldsymbol{y_n} &\sim P^{MVP}_{\theta(m)}(\boldsymbol{y} \mid \boldsymbol{x})\\
  \end{split}
\end{align}

We then perform majority voting and select the most frequent response as follows:
\begin{equation}
  \boldsymbol{y} \sim \#\argmax{\{i \mid y_i = \boldsymbol{y}\}}
\end{equation}

The proposed approach improves upon existing techniques in several ways.

First, MVP allows for considering several models trained on different datasets, which is especially
useful for complex problems or when a single model does not have sufficient knowledge to generate
proper responses. As opposed to existing strategies that query the same model multiple
times~\cite{arora2023ask}, we adopt an approach similar to that seen in the Random Forest, where the
final prediction depends on the majority vote of multiple learners~\cite{randomforests}. Despite the
individual models not being weak learners, utilizing a pool of models in the proposed manner may
also balance bias and variance and converge to an average obtained from multiple datasets. In other
words, a single LLM generates text after pre-training on some dataset \(D_1\), while MVP uses the
knowledge from the collection of datasets \(\{D_1, D_2, \dots, D_n\}\). While the datasets may
overlap, i.e., \(\bigcap_{i=1}^n D_i \neq \emptyset\) can be true, the increasing availability of
domain-specific datasets has facilitated the development of domain-aware LLMs. Besides, the
availability of general-purpose conversational datasets has also been
growing~\cite{henderson-etal-2019-repository}. Thus, it is not too difficult to find a set of models
where the overlap is not large~\cite{lewispretrained, gururanganpretrain}. Therefore, while the
importance of strategies utilizing a single model cannot be understated, it is vital to consider
approaches for improving performance that make use of multiple models.

Second, MVP facilitates inference on low-powered hardware (e.g., with limited GPU memory). In cases
where using very large (over 50 billion parameters) LLMs is infeasible due to cost or hardware
constraints, one can combine results from several smaller models and observe performance improvement
against any individual model.

Finally, since MVP is an aggregator of results, it is flexible, making the integration of
pre-existing prompting methods straightforward.


\section*{Dataset}

In this section, we introduce a new rare disease dataset. First, we would like to describe the
methods for obtaining the rare disease dataset, including generating the subset and the annotation
process. We used the recently released MIMIC-IV database, which is over five times larger than its
predecessor MIMIC-III and contains 331,794 de-identified discharge summaries, as the base for the
dataset. Figure~\ref{fig:dataset_pipeline} shows the overview pipeline.

\subsection*{Term Matching Augmented with Weak Supervision Rules}

We used SPARQL domain-specific language (DSL) for extracting rare disease terms from the Orphanet
Rare Disease Ontology (ORDO) version
4.2\footnote{\url{https://www.orphadata.com/ordo/}}
\cite{vasant2014ordo}. SPARQL queries were executed using the Python library
\texttt{rdflib}\footnote{\url{https://rdflib.readthedocs.io/en/stable/}}.
After obtaining a list of rare diseases, we performed simple term matching on the MIMIC-IV database.
Since the number of lookups was on the order of billions and the documents were not of small size,
we have not used Python. The rationale was two-fold: first, due to Global Interpreter Lock (GIL),
Python does not have the ability to perform proper multi-threading and second, it is a high-level
interpreted language, which would not satisfy the performance requirements needed for efficiently
completing the task at hand. Instead, we used the Rust\footnote{\url{https://www.rust-lang.org/}} programming language and the
\texttt{rayon}\footnote{\url{https://docs.rs/rayon/latest/rayon/}}
library for performing term-matching in parallel. This allowed us to create an inverted index of
terms mapped to the note identifiers.

After obtaining an inverted index, we performed further filtering by applying weak supervision rules
proven effective in the recent work on rare diseases~\cite{rdws}, by greatly improving the precision
while retaining the level of recall of a string-based matching method. Specifically, we removed rare
diseases whose term length was less than four (i.e., character count rule to filter out ambiguous
abbreviations) or those whose occurrences were more than 0.5\% (i.e., ``prevalence'' rule to filter
out common disease mentions which are not likely to be of a rare disease). Finally, we obtained a
subset from which we selected the four most frequent rare diseases, having a number of cases
sufficient for us to perform the few-shot evaluation on them. These four rare diseases are
Babesiosis, Giant Cell Arteritis, Graft Versus Host Disease, and Cryptogenic Organizing Pneumonia.

\begin{table}
  \centering
  \begin{tabular}{lcccc}
    \toprule
    \textbf{Disease} & \textbf{MIMIC-IV} & \textbf{Filtered} & \textbf{AR 1} & \textbf{AR 2}\\
    \midrule
    Babesiosis                       & 320      & 106       & 16   & 64\\
    Cryptogenic Organizing Pneumonia & 304      & 110       & 16   & 64\\
    Giant Cell Arteritis             & 406      & 115       & 16   & 64\\
    Graft Versus Host Disease        & 429      & 106       & 16   & 64\\
    \bottomrule
  \end{tabular}
  \caption{Number of documents per disease.}
  \label{tab:dataset}
\end{table}

\subsection*{Annotations}

The annotation process consisted of two rounds: an initial session to ensure a high IAA and a second
session for the final annotations. Two annotators with specialized knowledge in the biomedical field
annotated the two distinct batches of CNs with rare disease occurrences. The initial round had 64
CNs, while the second round had 256 CNs. If a patient whose CN was under consideration had a rare
disease, the patient would be labeled ``1'' and ``0'' otherwise. Cases where a patient's CN
demonstrated a family history of the disease, or suffered from that disease in the past (but not in
the present), were labeled as ``0''\footnote{To note that this is distinct from Dong et
al.~\cite{rdws} where \textit{past} rare diseases were also annotated as positive; we aim to
identify \textit{present} rare diseases.}. Our annotation guidelines outlined the annotation
process, accompanied by examples of positive and negative matches.

We used Cohen's kappa~\cite{kappa} for computing IAA:

\begin{equation}
  \kappa = \frac{p_o - p_e}{1 - p_e}
\end{equation}

where \(p_0\) is the probability of agreement on the label assigned to any sample, and \(p_e\) is
the hypothetical probability of chance agreement.

The Cohen's kappa score for the initial IAA assessment on 64 document annotations (the first round
of annotations) was 0.839. Such a high value suggested a near-perfect agreement after which, we
moved to the second round.

In the second round, 256 documents were annotated, which we used for evaluating the proposed
prompting method.

Finally, we ended up with 256 annotation documents providing information about which rare disease
occurred in a context and whether the person (whose discharge summary it is) suffers from the rare
disease present in the context. The final dataset statistics are shown in Table~\ref{tab:dataset}.
The dataset will be available to researchers who have signed MIMIC-IV Data Use Agreement (DUA).


\section*{Experiments}

\begin{figure*}[t!]
  \begin{center}
    \includegraphics[width=\linewidth]{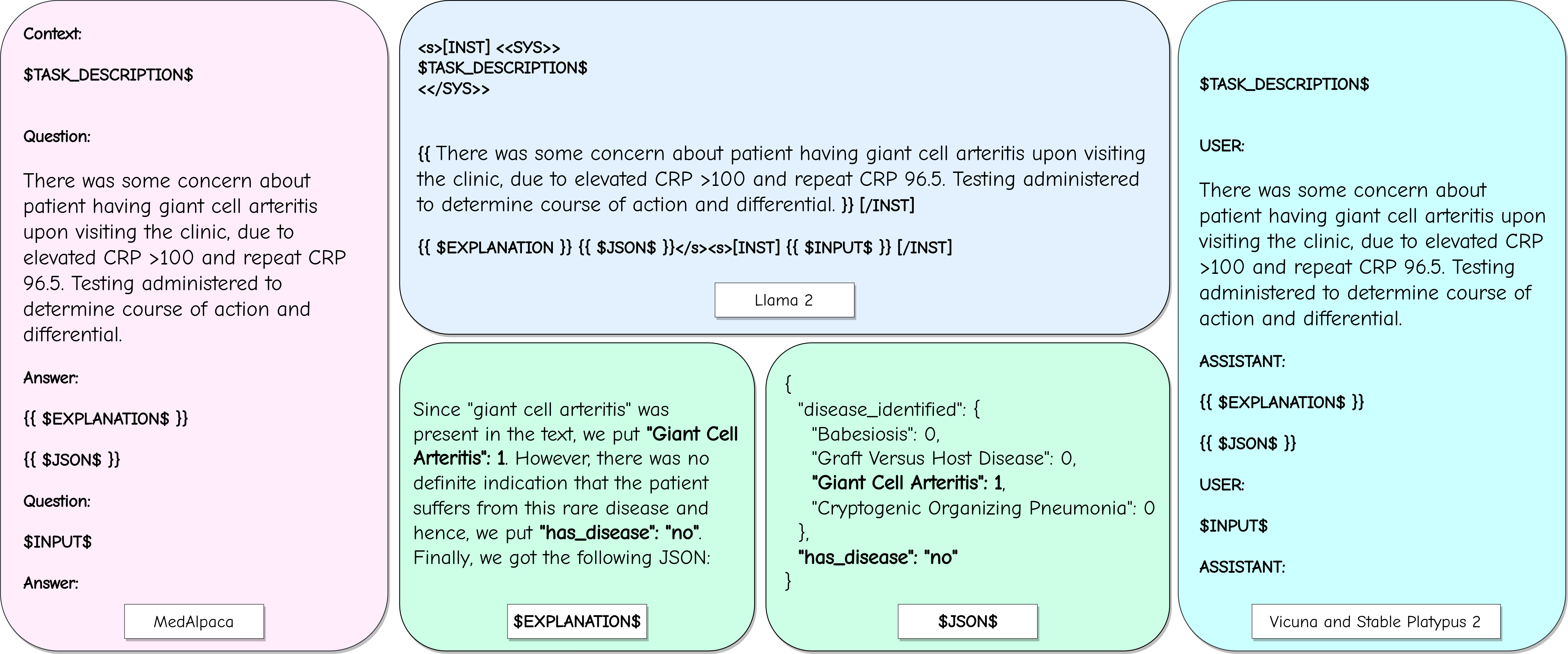}
  \end{center}
  \caption{CoT-Augmented Models-Vote Prompt Engineering}
  \label{fig:prompt_engineering}
\end{figure*}

The rare disease dataset we built was used for FSL experiments. For a more holistic evaluation, the
original 256 annotated documents were chunked into four subsets, containing CN substrings with 32,
64, 128, and 256 words (context size). It should be noted that every context window contained the
rare disease mention. The evaluation was performed using the following openly available models:
Llama~2~13B~\cite{llama2}, MedAlpaca~13B~\cite{medalpaca},
Stable~Platypus~2~13B~\cite{stableplatypus2}, and Vicuna~13B~\cite{vicuna}. We selected these LLMs
as they were some of the highest scoring models on the
Open~LLM~Leaderboard~\cite{openllmleaderboard}. MedAlpaca specifically was selected due to its
fine-tuning on medical question-answering. To compare MVP with SC prompting, Llama 2 with SC
prompting was also evaluated. For all models, the maximum number of tokens was set to 1,024.

\subsection*{Tasks}

As discussed before, we considered two tasks: rare disease identification and rare disease
classification.

Rare disease identification, while discussed separately from classification, can also be considered
a binary classification task. This task determines whether the model can infer that the person in
question has a particular disease. In this case, majority voting consisted of computing the number
of 0 and 1 votes per prompt. Since our experiments considered 4 models in total, if the sum was less
than 2, we counted it as no (0). Otherwise, we counted it as yes (1).

Rare disease classification assesses the model's ability to correctly identify rare diseases in the
given context. For this task, we had five classes: the original four rare diseases and the class
``other'' if the model prediction was not one of the four diseases. The disease was considered
correctly classified if it received the majority of votes. In situations where multiple labels were
predicted (i.e., had majority votes), as long as the correctly identified one was among them, we
still counted it as a correct prediction. The rationale is that having the same number of votes
makes a sample stand out and in practice would require careful examination.

In total, 256 CNs have been considered and for each of the tasks, we provided 32, 64, 128, and 256
word context windows for evaluation.

\subsection*{Prompt Engineering}

Proper calibration of prompts has demonstrated improvement in model
performance~\cite{zhao2021calibrate}. Similarly, using the same prompt format as in the pre-training
can improve model performance, and such prompt engineering approaches have gotten
popular~\cite{promptalike}. We followed the paradigm and designed prompts on a per-model basis.
Figure~\ref{fig:prompt_engineering} illustrates the method.

We used CoT in the prompt design. For MedAlpaca, the template included a task description as a
context, a question-and-answer example for instruction prompting, and the actual question in the
form of a CN context. Llama~2 prompt followed the same idea but used tags specific to its
pre-training. Finally, for both Vicuna and Stable Platypus 2, we utilized an approach similar to
MedAlpaca, as their pre-training prompts were also alike. Note that \texttt{\$EXPLANATION\$},
\texttt{\$JSON\$}, and \texttt{\$TASK\_DESCRIPTION\$} are all placeholders to be replaced by actual
text. For \texttt{\$EXPLANATION\$}, \texttt{\$JSON\$}, text is shown in
Figure~\ref{fig:prompt_engineering}. \texttt{\$TASK\_DESCRIPTION\$} describes a task and lists all
diseases to be identified, but has been omitted for brevity.

\subsection*{JSON-Augmented Prompts to Facilitate Model Evaluation}

Automated evaluation of generative LLMs can be challenging, as the output is mainly human language.
A typical solution is recruiting annotators, which can be expensive and time-consuming. Recently,
studies have shown that using formats, such as XML, can work well for model
evaluation~\cite{chen2023large}.

As shown in Figure~\ref{fig:prompt_engineering}, we used JSON format to represent a part of the
input prompt. Since the instruction output contained a parsable JSON string, generative LLMs
replicated the behavior, which reduced the need for human annotation and allowed for automatic
evaluation.

We also note that using JSON for prompt engineering is flexible and not limited to rare disease
identification. Furthermore, we can use JSON to model nested responses and graph-like structures as
we can describe graphs using adjacency lists.


\subsection*{Metrics}

For model evaluation, we used accuracy, precision, recall, and F-score. We also used paired t-tests
between the best and second-best models to verify that the mean difference between two sets of
observations is non-zero (i.e., there is a difference in model behavior). Note that t-tests were
performed directly on model outputs, and the threshold for statistical significance was 0.05. The
null hypothesis states that model prediction scores are not statistically significantly different,
and an alternative hypothesis states the opposite.


\section*{Results}

\begin{table}
  \centering
  \footnotesize
  \begin{tabular}{llll}
    \toprule
    \textbf{Model} & \textbf{Context} & \textbf{Identification (APRF)} & \textbf{Classification (APRF)}\\
    \midrule
    \textit{Llama 2}                              &           &    0.63,    0.68,    0.61,     0.59    &    0.77, \h{0.79},   0.62,     0.69\\
    \textit{MedAlpaca}                            &           &    0.48,    0.51,    0.50,     0.41    &    0.32,    0.62,    0.26,     0.29\\
    \textit{Stable Platypus 2}                    & 32 words  &    0.63,    0.63,    0.63,     0.63    &    0.74,    0.78,    0.59,     0.67\\
    \textit{Vicuna}                               &           &    0.60,    0.63,    0.61,     0.59    &    0.70,    0.70,    0.56,     0.61\\
    \textit{Llama 2 (Self-Consistency Prompting)} &           &    0.60, \h{0.72},   0.57,     0.50    & \h{0.84},   0.78, \h{0.67}, \h{0.72}\\
    \textit{Models-Vote Prompting}                &           & \h{0.66},   0.66, \h{0.65}, \h{0.65}   &    0.80,    0.76,    0.64,    0.69\\
    \midrule
    \textit{Llama 2}                              &           &    0.63,     0.67,     0.61,     0.58  &    0.74,    0.76,    0.59,    0.66\\
    \textit{MedAlpaca}                            &           &    0.47,     0.49,     0.50,     0.39  &    0.28,    0.63,    0.23,    0.24\\
    \textit{Stable Platypus 2}                    & 64 words  &    0.67,     0.67,     0.66,     0.66  &    0.76,    0.78,    0.61,    0.68\\
    \textit{Vicuna}                               &           &    0.62,     0.64,     0.63,     0.62  &    0.77,    0.75,    0.62,    0.67\\
    \textit{Llama 2 (Self-Consistency Prompting)} &           &    0.60,  \h{0.79},    0.57,     0.49  &    0.79, \h{0.79},   0.63, \h{0.70}\\
    \textit{Models-Vote Prompting}                &           & \h{0.70},    0.72,  \h{0.69}, \h{0.69} & \h{0.81},   0.75, \h{0.65},   0.69\\
    \midrule
    \textit{Llama 2}                              &           & \h{0.62},    0.65,   0.60,     0.58   &    0.70, \h{0.79},      0.56,    0.66\\
    \textit{MedAlpaca}                            &           &    0.48,     0.53,    0.51,     0.40   &    0.25,    0.59,      0.20,    0.17\\
    \textit{Stable Platypus 2}                    & 128 words &    0.61,     0.61, \h{0.61}, \h{0.60}  &    0.71, \h{0.79},     0.57,    0.66\\
    \textit{Vicuna}                               &           &    0.59,     0.61,    0.60,     0.58   &    0.70,    0.75,      0.56,    0.63\\
    \textit{Llama 2 (Self-Consistency Prompting)} &           &    0.60,  \h{0.68},   0.57,     0.51   & \h{0.75}, \h{0.79}, \h{0.60}, \h{0.68}\\
    \textit{Models-Vote Prompting}                &           & \h{0.62},    0.62, \h{0.61}, \h{0.60}  & \h{0.75},   0.77,   \h{0.60},   0.67\\
    \midrule
    \textit{Llama 2}                              &           &    0.67,    0.67,     0.66,     0.66   &    0.61,    0.79,    0.48,     0.59\\
    \textit{MedAlpaca}                            &           &    0.47,    0.51,     0.50,     0.35   &    0.19,    0.32,    0.15,     0.10\\
    \textit{Stable Platypus 2}                    & 256 words &    0.61,    0.61,     0.60,     0.60   & \h{0.66},   0.76, \h{0.53}, \h{0.61}\\
    \textit{Vicuna}                               &           &    0.67, \h{0.71}, \h{0.68},    0.66   &    0.52,    0.70,    0.41,     0.50\\
    \textit{Llama 2 (Self-Consistency Prompting)} &           &    0.63,    0.65,     0.62,     0.60   &    0.62, \h{0.80},   0.50,     0.60\\
    \textit{Models-Vote Prompting}                &           & \h{0.68},   0.67,     0.67,  \h{0.67}  &    0.61,    0.73,    0.49,     0.56\\
    \bottomrule
  \end{tabular}
  \caption{Experimental Results on Identification and Classification of Rare Diseases}
  \label{tab:results}
\end{table}

Table~\ref{tab:results} shows the experimental results. Note that APRF is a 4-tuple that stands for
Accuracy, Precision, Recall, and F-score, respectively. Moving forward, we will use the APRF acronym
to discuss performance. The best scores for a given context will be highlighted as bold and
underlined.

\subsection*{Rare Disease Identification}

For 32-word context windows, MVP performed the best, with the APRF of (0.66,~0.66,~0.65,~0.65).
Llama~2 came next with the APRF of (0.63,~0.68,~0.61,~0.59). Interestingly, Llama 2 (SC Prompting)
outperformed MVP in precision but underperformed in all other metrics. Stable Platypus 2 and Vicuna
had comparable performance, with MedAlpaca showing the worst performance. The paired t-test for
difference on MVP and Llama 2 outputs resulted in a statistically significant p-value of
approximately \(7.437\mathrm{e}{-9}\).

MVP also showed the best performance in the case of 64-word contexts, with the APRF of
(0.70,~0.72,~0.69,~0.69). Stable Platypus 2 had the second-best performance of
(0.67,~0.67,~0.66,~0.66). Llama 2 (SC Prompting) had the best overall precision score but did not
perform as well in other metrics. Llama 2 and Vicuna had comparable performance, while MedAlpaca had
the worst APRF values. The paired t-test on MVP and Stable Platypus 2 resulted in a statistically
significant p-value of approximately \(1.600\mathrm{e}{-2}\).

As for 128-word context windows, MVP and Llama 2 showed the best performance, with the APRF values
of (0.62,~0.62,~0.61,~0.60) and (0.62,~0.65,~0.60,~0.58), respectively. Llama 2 (SC Prompting)
showed the best precision, but underperformed in other metrics. Stable Platypus 2 and Vicuna
performed similarly, with MedAlpaca performing the worst. The paired t-test on MVP and Stable
Platypus 2 gave a statistically significant p-value of approximately \(2.928\mathrm{e}{-7}\).

For 256-word context windows, MVP and Vicuna performed similarly, with APRF values of
(0.68,~0.67,~0.67,~0.67) and (0.67,~0.71,~0.68,~0.66), respectively. Llama 2, Llama 2 (SC Prompting),
and Stable Platypus 2 had comparable performance, with MedAlpaca showing the worst
performance. The p-value for the paired t-test on MVP and Vicuna was \(7.979\mathrm{e}{-13}\),
demonstrating statistical significance in the mean difference of outputs.

Overall, MVP showed the best performance across all benchmarks. The difference in outputs of MVP and
second-best models was verified by the paired t-test for difference, with p-values always being less
than the cutoff value of 0.05. MVP also outperformed Llama 2 (SC Prompting) across all context
sizes.

\subsection*{Rare Disease Classification}

For the 32-word context windows, Llama 2 (SC Prompting) performed the best with the APRF of
(0.84, 0.78, 0.67, 0.72). MVP was the second-best approach and had the APRF of
(0.80, 0.76, 0.64, 0.69). Llama 2, Stable Platypus 2, and Vicuna performed similarly, with MedAlpaca
performing the worst. The paired t-test between Llama 2 (SC Prompting) and MVP gave a statistically
significant p-value of approximately \(3.370\mathrm{e}{-1}\).

As for the 64-word context windows, MVP performed the best with the APRF values of
(0.81, 0.75, 0.65, 0.69). Llama 2 (SC Prompting) marginally underperformed with the APRF of
(0.79, 0.79, 0.63, 0.70). Llama 2, Stable Platypus 2, and Vicuna had similar performance. MedAlpaca
showed the worst performance. The paired t-test on the outputs of MVP and Llama 2 (SC Prompting)
resulted in the statistically significant p-value of \(1.570e-5\).

In 128-word context experiments, Llama 2 (SC Prompting) and MVP performed the best, with the APRF
values of (0.75,~0.79,~0.60,~0.68) and (0.75,~0.77,~0.60,~0.67), respectively. MVP performed
slightly underperformed. Llama 2, Stable Platypus 2, and Vicuna had similar performance, with
MedAlpaca performing the worst. The p-value value for Llama 2 (SC Prompting) and MVP was
approximately \(5.251\mathrm{e}{-3}\), which is not statistically significant.

Finally, for 256-word experiments, Stable Platypus 2 and Llama 2 (SC Prompting) were first and
second best models, with the APRF scores of (0.66,~0.76,~0.53,~0.61) and (0.62,~0.80,~0.50,~0.60),
respectively. Llama 2, Vicuna, and MVP had similar performance, and MedAlpaca performed the worst.
The p-value for the t-test between Stable Platypus 2 and Llama 2 (SC Prompting) was approximately
\(1.260\mathrm{e}{-2}\), meaning that the mean difference between model outputs was not
statistically significant.

In rare disease classification results, MVP and Llama 2 (SC Prompting) showed the best overall
results. However, unlike rare disease identification, in the case of 256-word context windows, both
Stable Platypus 2 and Llama 2 (SC Prompting) marginally outperformed MVP.

\subsection*{JSON Compliance in Model Evaluation}

Table \ref{tab:json} shows the number of model responses that were not in partial or complete
JSON-compliant format. These leftover entries were human-annotated. This was a substantially faster
method, as approximately 85.9\% of the 4096 results did not require manual annotation. As shown in
the first column of the table, MedAlpaca had the worst compliance rate for JSON format, with 197
errors across 4 context sizes. Llama 2 on the other hand performed very well, with only 33 errors.

\begin{table}
  \centering
  \begin{tabular}{lcc}
    \toprule
    \textbf{Model} & \textbf{No JSON} & \textbf{Compliance}\\
    \midrule
    Llama 2            & 33      & 96.8\%\\
    MedAlpaca          & 197     & 80.8\%\\
    Stable Platypus 2  & 185     & 82.0\%\\
    Vicuna             & 162     & 84.2\%\\
    \midrule
    Overall JSON       & 577     & 85.9\%\\
    \bottomrule
  \end{tabular}
  \caption{JSON Compliance by Model}
  \label{tab:json}
\end{table}

\section*{Ablation Study}

\begin{figure*}[h!]
  \begin{center}
    \includegraphics[width=\linewidth]{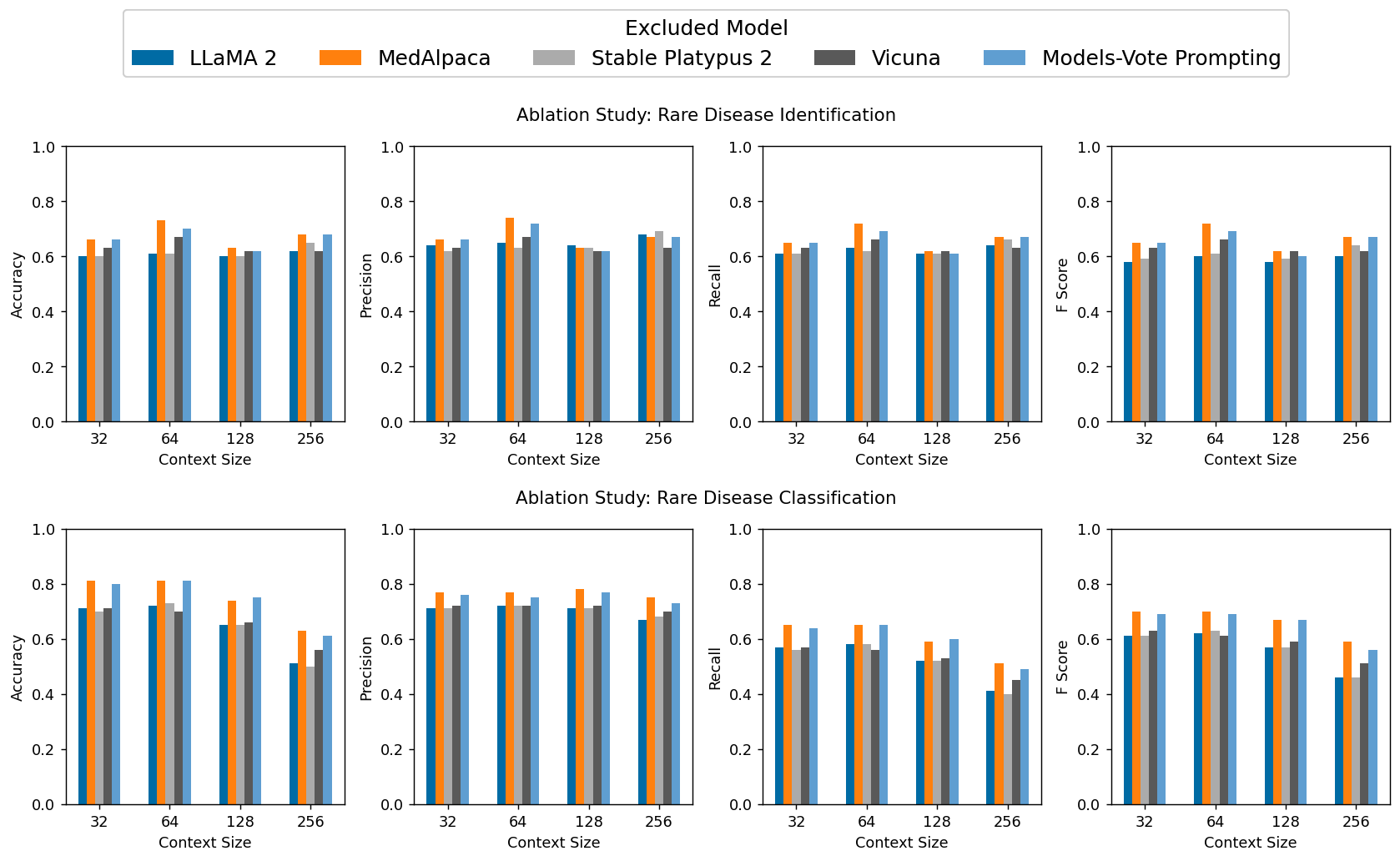}
  \end{center}
  \caption{Ablation Study Results}
  \label{fig:ablation_study}
\end{figure*}

We conducted ablation study to remove individual models from MVP and examined the model performance.
Figure~\ref{fig:ablation_study} shows the results of the ablation study. Note that Models-Vote
Prompting label denotes the original MVP performance with all models, with no model model excluded.

Overall, removing a model from the MVP ensemble decreases performance across APRF. However, in the
case of MedAlpaca, removing the model causes MVP to have a similar or slightly improved performance
to the original four-model ensemble. An explanation for the change in performance could be
MedAlpaca's underperformance on the given tasks (relative to the other models). This behavior is
consistent across the identification and classification tasks for all tested context sizes.
Furthermore, for the rare disease identification task at 128-word context size, removing individual
models does not result in significant performance degradation (i.e., the resulting performance is
comparable to that of the original MVP). Such behavior may be due to the models exhibiting similar
performance at the 128-word context and Vicuna not performing as well as it does for other context
sizes.

Both the number of models and models themselves are hyperparameters, and we hypothesize that
determining optimal values may be domain and task-dependent. In the following section, we also note
that this could be a potential future research direction.


\section*{Limitations and Future Work}

There is much to be explored and expanded upon beyond what we presented in the paper. First, using
other LLMs for model evaluation, such as Falcon~\cite{falcon40b} whose training data differs from
that of models used in the paper, could be interesting. Second, performing the same tasks using
smaller LLMs (e.g., 7 billion parameter models) may show promising results. Third, increasing the
number of models used in MVP can be another research direction. Fourth, the criteria for the
selection of models for MVP needs further study. Fifth, using MVP for other domains or tasks can
also be an option. The novel rare disease dataset can help define new evaluation tasks, and
incorporating different prompting approaches may help improve MVP performance. Note that increasing
context size decreased model performance (in both tasks). This behavior may have been caused by
increased ambiguity, as a larger context window often contains more medical terms, which increases
complexity and reduces performance. However, a thorough study is needed to examine and better
explain the behavior. Finally, using MVP with prompts incorporating JSON for depicting other
response structures (e.g., nested relationships, graph-like structures, etc.) can also be a
potential future avenue for exploration.


\section*{Conclusion}

We proposed Models-Vote Prompting (MVP), a prompting approach that, in addition to its promising
performance on the novel rare disease dataset, introduces a new perspective to rare disease
identification and classification tasks. Through our experiments, we evaluated and quantified the
extent of the performance improvements achieved by our method. To support the hypothesis presented
in the paper, we conducted statistical hypothesis tests that demonstrated statistical significance
across experiments. Furthermore, we conducted an ablation study to examine the behavior of MVP after
excluding individual models. We also explored the feasibility of using JSON-augmented prompts, which
proved effective and reduced the need for manual, human annotation of the LLM-generated results.


\section*{Acknowledgements}

This work was supported by the National Institutes of Health under award number U24TR004111 and
R01LM014306. The content is solely the responsibility of the authors and does not necessarily
represent the official views of the National Institutes of Health.


\section*{Acknowledgements}

This work was supported by the National Institutes of Health under award number U24TR004111 and
R01LM014306. The content is solely the responsibility of the authors and does not necessarily
represent the official views of the National Institutes of Health.


\section*{Ethics Declarations}

\subsection*{Competing Interests}

Y.W. consults for Pfizer Inc. and has ownership/equity interests in BonafideNLP, LLC. All other
authors declare no competing interests.


\section*{Data Availability}

The dataset is reproducible to those who signed MIMIC-IV DUA and therefore, have access to MIMIC-IV.


\section*{Code Availability}

\url{https://github.com/PittNAIL/llms-vote}


\printbibliography


\end{document}